\documentclass[runningheads]{llncs}
\usepackage[T1]{fontenc}
\usepackage{multirow}
\usepackage{xurl}
\usepackage{subcaption}
\usepackage{array}

\usepackage{tikz}
\usetikzlibrary{shapes, arrows, positioning}

\usepackage{listings}
\usepackage{fancyvrb}
\usepackage{pmboxdraw}
\makeatletter
\newcommand\notsotiny{\@setfontsize\notsotiny\@vipt\@viipt}
\makeatother

\begin{document}
\title{LLMs for Text-Based Exploration and Navigation under Partial Observability}
\titlerunning{LLMs for Text-Based Exploration \& Navigation under Partial Observability}

\author{
  Stephan Sandfuchs\inst{1}\orcidID{0000-0001-7621-7191} \and
  Maximilian Melchert\inst{1}\orcidID{0009-0002-8412-2488} \and
  Jörg Frochte\inst{1}\orcidID{0000-0002-5908-5649}
}
\authorrunning{S. Sandfuchs, M. Melchert \& J. Frochte}

\institute{
  AKIS -- Interdisciplinary Institute for Applied AI and Data Science Ruhr,\\
  Bochum University of Applied Sciences, Bochum, Germany\\
  \email{\{stephan.sandfuchs,maximilian.melchert,joerg.frochte\}@hs-bochum.de}
}

\maketitle

\begin{abstract}
Exploration and goal-directed navigation in unknown layouts are central to inspection, logistics, and search-and-rescue.
We ask whether large language models (LLMs) can function as \emph{text-only} controllers under partial observability -- without code execution, tools, or program synthesis.
We introduce a reproducible benchmark with oracle localisation in fixed ASCII gridworlds: each step reveals only a local $5\times5$ window around the agent and the model must select one of \texttt{UP/RIGHT/DOWN/LEFT}.
Nine contemporary LLMs ranging from open/proprietary, dense / Mixture of Experts and instruction- vs.
reasoning-tuned are evaluated on two tasks across three layouts of increasing difficulty: \emph{Exploration} (maximising revealed cells) and \emph{Navigation} (reach the goal on the shortest path).
The experimental results are evaluated on quantitative metrics including  \emph{success rate}, \emph{efficiency} such as normalised coverage and \emph{path length} vs. oracle as well as qualitative analysis.
Reasoning-tuned models reliably complete navigation across all layouts, yet remain less efficient than oracle paths.
Few-shot demonstrations in the prompt chiefly help these Reasoning-tuned models by reducing invalid moves and shortening paths, while classic dense instruction models remain inconsistent. 
We observe characteristic action priors (UP/RIGHT) that can induce looping under partial observability.
Overall, training regimen and test-time deliberation predict control ability better than raw parameter count.
These findings suggest lightweight hybridisation with classical online planners as a practical route to deployable partial map systems.
\end{abstract}

\section{Introduction}
\label{sec:Introduction}
Exploration and goal-directed navigation in previously unknown layouts are core requirements in inspection, logistics, and search-and-rescue.
While classical methods excel once a reliable map exists, many deployments must discover structure on the fly under tight resource constraints.
In parallel, large language models (LLMs) have shown abilities in structured decision making via text interfaces, motivating the question: 
to what extent can LLMs act as controllers when the environment is only partially observed?

We study a deliberately simple yet application-relevant setting: a discrete ASCII gridworld with \emph{oracle localisation} -- the agent's coordinates are always known --while unobserved cells are denoted by \texttt{?}.
This is not SLAM: the uncertainty is in the map, not in pose.
We evaluate two tasks that capture common operational needs: (i) \emph{Exploration}, where the agent aims to reveal the environment under a step budget; and (ii) \emph{Obstacle-aware navigation}, where it must reach a specified target while avoiding obstacles on an otherwise unknown layout.

Our interface casts the problem as myopic, one-step action selection.
At each time $t$, the model receives the current map state and the immediately preceding state, then outputs a single action from a constrained schema \texttt{UP}, \texttt{DOWN}, \texttt{LEFT}, \texttt{RIGHT}.
The simulator executes the action, updates the map, and returns the new state along with simple feedback for invalid moves.
\emph{Program synthesis and tool use are disallowed}: models must decide actions directly in natural language; no code generation, external tools, or classical planners may be invoked.
This isolates planning-like behaviour and minimal memory from perception and state estimation, yielding a deterministic, reproducible testbed.

We ask whether LLMs -- without code execution or external tools -- can reliably perform stepwise action selection under partial maps.
We hypothesise that few-shot demonstrations help initial exploration, but do not close the efficiency gap to classical re-planners as environments grow.

We conduct a comparative study across nine contemporary LLMs (two proprietary, seven open-weight) and three environments of increasing difficulty, each evaluated on \emph{Exploration} and \emph{Navigation}.
We report engineering-oriented metrics -- success rate and efficiency such as normalised coverage and \emph{path length} -- and complement them with selected case studies that illustrate characteristic behaviours and failure modes under partial observability.

Our main contributions are as follows:
\begin{enumerate}
\item We introduce a benchmark framework for stepwise action selection with LLMs in partially observable, oracle-localised gridworlds, enforcing strict no-code/no-tool constraints and evaluating with deployability-focused metrics.
\item We present a broad, systematic evaluation of nine LLMs -- spanning proprietary and open-weight models -- across a spectrum of tasks and environments.
\item We provide case studies and analyses that characterise the strengths, limitations, and failure modes of LLMs under these realistic constraints, offering design guidance for practitioners.
\end{enumerate}

The remainder of this paper is organised as follows.
Section~\ref{sec:sota} reviews related work.
Section~\ref{sec:Experimental:Setup} specifies the tasks and environments (fixed $19\times21$ grids with $5\times5$ reveal) as well as the model interface and prompting regimes (zero- and $n$-shot).
Section~\ref{Results:and:Discussion} presents the evaluation metrics, reports results, and discusses observed behaviours and failure modes.
Section~\ref{sec:Conclusion:and:Future:Prospects} concludes with limitations and future prospects.

\section{State of the Art}
\label{sec:sota}
Decades of work in robotics have addressed sequential decision-making under partial observability with mature stacks for localisation, mapping (SLAM), planning, and control (see, e.g., \cite{cadena16slam,bailey06slam,thrun05prob,sandfuchs22periodic}).
In our case, localisation is known and therefore omitted; the canonical engineering solution under partial maps is to replan online on the revealed occupancy grid -- frontier-based exploration \cite{Yamauchi1997} and incremental shortest-path search such as D* Lite \cite{Koenig2002}.
These methods calibrate path and coverage efficiency in practice, yet they rely on algorithmic solvers that conflict with our strict no-tool constraint.
We therefore cite them as context and evaluate LLMs only between random and oracle bounds to isolate text-only control.

Text-only gridworlds offer a controlled way to study planning-like behaviour under partial observability while abstracting perception and SLAM.
Early work showed that LLMs can be prompted to act as high-level planners or next-action selectors in symbolic or textual environments, but reliability degrades on longer horizons and more complex layouts \cite{huang22,yao22react}.
Our evaluation follows this line yet imposes \emph{oracle localisation} (pose known; map unknown), \emph{stepwise control} (one action per turn), and a strict \emph{no-code constraint} (no program synthesis or external tools).
The map is fixed to $19\times21$ cells, and each move reveals a local $5\times5$ neighbourhood around the agent; these choices isolate decision-making under limited, structured observability and avoid confounds from perception.

\emph{Representation matters}.
The textual format used to encode spatial information (e.g., Cartesian coordinates versus topographic ASCII layouts) can substantially affect success rates and path efficiency in grid navigation, and internal activations contain units correlated with spatial features \cite{martorell25text2space}.
Such findings indicate that input structure and prompt design can modulate behaviour but do not by themselves guarantee robust long-horizon control.

A central question is generalisation beyond training distributions.
Evidence from textualised gridworlds indicates that next-token methods -- and even chain-of-thought fine-tuning -- often fail to extrapolate to larger, unseen mazes without specialised training that induces structured ``cognitive maps''; simple prompting proved insufficient in controlled tests \cite{kim24textgrid}.
Related results on compositional and systematic generalisation (e.g., SCAN and COGS) likewise report near-perfect in-distribution accuracy but sharp drops on held-out combinations, highlighting limits of current neural learners under distribution shift \cite{LakeBaroni2018,kim2020cogs}.
Combined with the representation sensitivity above \cite{martorell25text2space}, this argues that observed successes may hinge on priors about the input format rather than on algorithmic planning skill.

In parallel, a strand of work augments LLMs with algorithms or lets them write and execute code to plan -- from tree / algorithm / program-of-thoughts prompting to world-models-as-code.
These hybrid systems report strong gains on grid tasks and long-horizon planning \cite{yao23tot,chen23pot,tang24worldcoder,li2025gridroute}, but they differ fundamentally from our evaluation: we prohibit tool use, program synthesis, and external solvers to isolate text-conditioned action selection.
We therefore cite hybrid work as context rather than as baselines.

Robotics also leverages learning for decision-making at the behaviour layer: learning-based arbitration among behaviour primitives can outperform hand-tuned selection policies \cite{tousside20behavsel}.
Connections between language and action often cast LLMs as high-level planners while low-level control remains classical; abstractions akin to grids or ``chessboards'' can serve as interfaces between the two \cite{ahn22saycan}.
Our study is purely textual and simulator-based but is motivated by the same goal: stress-test planning-like behaviour when only a partial map is observable and must be revealed online.

A practical challenge for sequential control is how much history to provide.
Although a full trajectory seems desirable, long-context studies consistently show that LLMs struggle to reliably use information buried in the middle of long inputs (``lost in the middle'') and degrade with distractors as contexts grow \cite{liu23lostmiddle,hsieh24ruler,longbench23}.
Industry reports describe a related phenomenon termed context rot, in which performance gradually deteriorates as stale content accumulates in the prompt \cite{chroma24report}.
In our domain, appending many prior ASCII maps introduces redundant snapshots that act as distractors; empirically this leads to confusion and looping.
Consequently, our interface intentionally restricts inputs to the current and the immediately previous map state with a concise action schema, trading trajectory completeness for robustness under partial observability.

Finally, our prompting regimes deliberately use zero-shot and $n$-shot demonstrations to bracket typical usage without fine-tuning.
Zero- and few-shot in-context learning is well-established \cite{brown20gpt3}, while chain-of-thought and zero-shot CoT prompting can improve stepwise reasoning but remain sensitive to formatting and task structure \cite{wei22cot,kojima22zshot}.
We report results under both regimes to separate sensitivity to demonstrations from the underlying ability to plan with limited, noisy state information.

Overall, the literature indicates that prompt-only stepwise control can succeed on modest text-encoded navigation but struggles with scaling, extrapolation, and partial observability \cite{huang22,yao22react,kim24textgrid}; that hybrid/code-driven methods can overcome some limits albeit outside our scope \cite{yao23tot,chen23pot,tang24worldcoder,li2025gridroute}; and that long contexts are brittle, so compact state representations mitigate the adverse effects of stale content \cite{liu23lostmiddle,hsieh24ruler,longbench23,chroma24report}.
These observations motivate our benchmark design and the emphasis on engineering-oriented metrics that capture deployability.

\section{Experimental Setup}
\label{sec:Experimental:Setup}
\begin{figure}[htbp]
    \centering
    \begin{tikzpicture}[
    node distance=3.5cm,
    auto
  ]

    \node (environment) [draw, rectangle, minimum height=1cm, minimum width=1.5cm, align=center] {Gridworld \\ Environment};
    \node (agent) [draw, rectangle, right=of environment, minimum height=1cm, minimum width=1.5cm, align=center] {Agent};
    \node (llm) [draw, rectangle, right=of agent, minimum height=1cm, minimum width=1.5cm, align=center] {LLM};

    \draw[->, line width=1.2pt] ([yshift=4pt]environment.east) -- ([yshift=4pt]agent.west)
        node[midway,above=2pt] {\scriptsize Observation}
        node[midway,above=12pt] {\scriptsize \textbf{(1)}};
    \draw[<-, line width=1.2pt] ([yshift=-4pt]environment.east) -- ([yshift=-4pt]agent.west)
        node[midway,below=2pt] {\scriptsize Action}
        node[midway,below=12pt] {\scriptsize \textbf{(4)}};

    \draw[->, line width=1.2pt] ([yshift=4pt]agent.east) -- ([yshift=4pt]llm.west)
        node[midway,above=2pt] {\scriptsize Prompt + Observation}
        node[midway,above=12pt] {\scriptsize \textbf{(2)}};
    \draw[<-, line width=1.2pt] ([yshift=-4pt]agent.east) -- ([yshift=-4pt]llm.west)
        node[midway,below=2pt] {\scriptsize Directional Decision}
        node[midway,below=12pt] {\scriptsize \textbf{(3)}};

\end{tikzpicture}
    \caption{
        Decision loop of an LLM-driven gridworld agent 
        (1) Environment: Observation + action evaluation (reward \& success signals); Agent: prompt generation, stop condition monitoring  
        (2) Agent: Structured prompt to LLM (system message, last action, observation)  
        (3) LLM: Generated decision of movement direction
        (4) Agent: Prompt validation, action forwarding to environment 
    }
    \label{fig:Decision:Loop}
\end{figure}
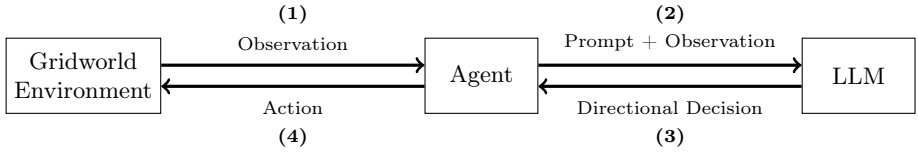

Figure \ref{fig:Decision:Loop} outlines our experimental setup: a simulated robot in a gridworld environment is controlled by LLM-generated action proposals, which an agent validates and executes.
In our study, the LLM component is instantiated by a broad set of models, with evaluation focusing on instruction-tuned and reasoning-oriented variants.

The gridworld environment utilises fixed $19\times21$ ASCII grids with a solid outer wall.
Walls are denoted \texttt{W}, free cells '\texttt{.}' or '\texttt{ }', the simulated robots position is \texttt{R}, and the goal is \texttt{g}.
The dynamics are fully deterministic.
The LLM generates a \{\texttt{DOWN},\texttt{RIGHT},\texttt{UP},\texttt{LEFT}\} action each step.
Attempts to move into a wall leave the simulated robot in place and are returned as invalid.
The environment is \emph{partially observed}: after every action the agent receives the current gridworld as an observation with unknown cells rendered as \texttt{?}.
Thus, each move reveals a $5\times5$ window centred on the robot (Chebyshev radius~2) and the already revealed cells persist.
Oracle localisation holds throughout (true coordinates are known).

The agent has to solve two primary robotics-inspired tasks: exploration and navigation.
In the exploration task the agent has to reveal the entire grid, i.e., reveal all \texttt{?} cells by visiting sufficient locations within $T_{\max}$ steps.
This task is terminates when either the whole grid is revealed or $T_{\max}$ is reached.
The metric used to compare the models performance in this task is the normalised exploration coverage, i.e. the number of revealed non-\texttt{?} cells divided by all cells in the grid.
This task isolates exploration under local sensing with no privileged global planner.
In the navigation task the agent has to reach a target position \texttt{g} from its start position \texttt{R} while the grid is initially unknown, except for the position of \texttt{g} and a $5\times5$ window centred on its position.
This task is either successful when \texttt{g} is reached or unsuccessful after $T_{\max}$ steps.
As a metric the success rate of multiple attempts and the average action count of successful attempts are utilized.
The requirement is to \emph{minimise} path length under partial information.
However, the exact shortest paths are not enforced at evaluation time.
For this setup we chose $T_{\max}=400=19\times21+1$ for both navigation and exploration task, because it provides roughly one action per cell, which is a conservative upper bound that avoids truncating near worst-case instances in both tasks.

\begin{figure}[tbp]
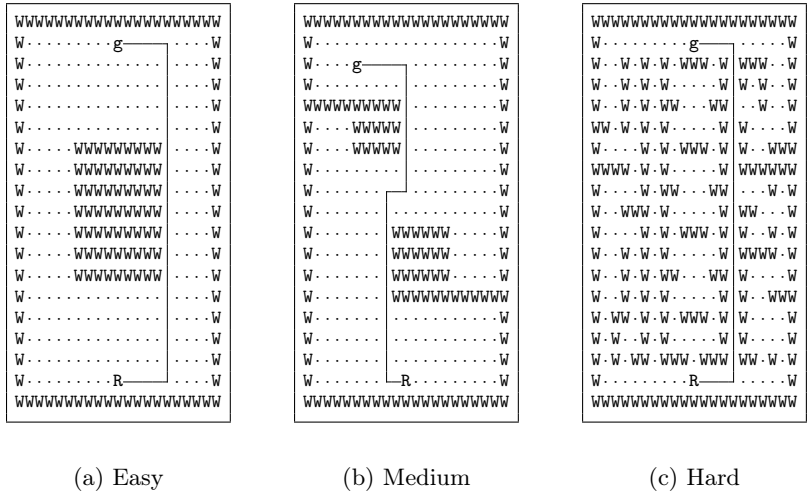

    \centering
    \begin{subfigure}{0.245\textwidth}
        \centering
\begin{Verbatim}
WWWWWWWWWWWWWWWWWWWWW
W·········g────┐····W
W··············│····W
W··············│····W
W··············│····W
W··············│····W
W·····WWWWWWWWW│····W
W·····WWWWWWWWW│····W
W·····WWWWWWWWW│····W
W·····WWWWWWWWW│····W
W·····WWWWWWWWW│····W
W·····WWWWWWWWW│····W
W·····WWWWWWWWW│····W
W··············│····W
W··············│····W
W··············│····W
W··············│····W
W·········R────┘····W
WWWWWWWWWWWWWWWWWWWWW
\end{Verbatim}
        \caption{Easy}
    \end{subfigure}
    \hspace{0.06\textwidth}%
    \begin{subfigure}{0.245\textwidth}
        \centering
\begin{Verbatim}
WWWWWWWWWWWWWWWWWWWWW
W···················W
W····g────┐·········W
W·········│·········W
WWWWWWWWWW│·········W
W····WWWWW│·········W
W····WWWWW│·········W
W·········│·········W
W·······┌─┘·········W
W·······│···········W
W·······│WWWWWW·····W
W·······│WWWWWW·····W
W·······│WWWWWW·····W
W·······│WWWWWWWWWWWW
W·······│···········W
W·······│···········W
W·······│···········W
W·······└─R·········W
WWWWWWWWWWWWWWWWWWWWW
\end{Verbatim}
        \caption{Medium}
    \end{subfigure}
    \hspace{0.06\textwidth}%
    \begin{subfigure}{0.245\textwidth}
        \centering
\begin{Verbatim}
WWWWWWWWWWWWWWWWWWWWW
W·········g───┐·····W
W··W·W·W·WWW·W│WWW··W
W··W·W·W·····W│W·W··W
W··W·W·WW···WW│··W··W
WW·W·W·W·····W│W····W
W····W·W·WWW·W│W··WWW
WWWW·W·W·····W│WWWWWW
W····W·WW···WW│···W·W
W··WWW·W·····W│WW···W
W····W·W·WWW·W│W··W·W
W··W·W·W·····W│WWWW·W
W··W·W·WW···WW│W····W
W··W·W·W·····W│W··WWW
W·WW·W·W·WWW·W│W····W
W·W··W·W·····W│W····W
W·W·WW·WWW·WWW│WW·W·W
W·········R───┘·····W
WWWWWWWWWWWWWWWWWWWWW
\end{Verbatim}
        \caption{Hard}
    \end{subfigure}
    \caption{   
        Full view of the used gridworlds with the oracle baseline for shortest navigation path with a length of (a) 26, (b) 25 and (c) 24. At the start of each attempt it is obscured with \texttt{?} except for a $5\times5$ window centred on \texttt{R} and \texttt{g} 
    }
    \label{fig:Gridworlds:Full:FOV}
\end{figure}

Figure~\ref{fig:Gridworlds:Full:FOV} illustrates the three layouts; all share the same outer frame but differ in internal structure.
The first environment \emph{Easy} contains a single barrier and the free space is simply connected (no cul-de-sacs).
The optimal route to \texttt{g} wraps around the barrier on either the left or the right.
\emph{Navigation} mainly tests whether the agent abandons an initially blocked heading and commits to circumnavigation.
The second environment \emph{Medium} contains an S-shaped channel formed by two obstacle regions.
Several shallow concavities (``bays'') branch off the main corridor and act as \emph{honeypots}: locally plausible but ultimately fruitless detours under a $5\times5$ view.
\emph{Exploration} requires systematic wall-following that does not over-spend in bays while \emph{navigation} penalises myopic ``peek into every recess'' behaviour and rewards early commitment along the main channel.
Finally the third environment \emph{Hard} is a near-symmetrical maze with multiple branches and cul-de-sacs.
From \texttt{R}, heading upwards leads to a slightly longer route with two lateral routes (left/right) providing the shortest solutions of similar length.
\emph{Exploration} demands deliberate backtracking to clear dead ends and \emph{navigation} stresses long-horizon choice under ambiguity -- early bias to ``go up'' yields large sub-optimality, whereas committing left or right yields compact solutions.

Taken together, the three layouts form a controlled difficulty ramp while keeping perceptual and actuation demands constant. They increase the task complexity progressively with the intent of demonstrating the performance degradation prompt-only control can exhibit as global structure increases.
\emph{Easy} emphasises \emph{unknown but simple} geometry (one global detour).
\emph{Medium} introduces \emph{local traps} that tempt greedy explorers.
\emph{Hard} adds \emph{global branching and cul-de-sacs}, requiring the agent to remember and revise choices as new evidence appears within the $5\times5$ reveal.
This separation lets us attribute failure modes to exploration versus long-horizon navigation rather than to sensing or stochastic.

For the LLM component we evaluate a representative mix of closed- and open-weight transformer LLMs spanning both dense and sparse architectures, including Mixture-of-Experts (MoE) models -- sparsely activated layers that maintain very large total capacity by routing each token through only a small subset of experts -- and models explicitly optimised for multi-step reasoning.
Throughout, ``Instruct'' denotes instruction-tuned checkpoints (supervised instruction data, typically followed by RLHF); we deliberately avoid chat-specialised endpoints.
Concretely, on the closed-weight side we use GPT-5 (reasoning, API-served flagship) and its smaller sibling GPT-5-mini (same family, lower latency/cost).
Our open-weight set comprises GPT-OSS-120B (reasoning-aligned MoE; Apache-2.0 open weights), Gemma-3-27B-Instruct (instruction-tuned Gemma-3 checkpoint), Qwen3-235B-A22B (MoE \emph{base} model with a switchable ``thinking/non-thinking'' mode rather than an instruct variant), Qwen3-QwQ-32B (Qwen's reasoning- specialised 32B release), DeepSeek-R1-0528 (an RL-driven reasoning model; we use the May-28 checkpoint), Llama-3.3-70B-Instruct (instruction-tuned, open-weight baseline), and Mistral-Large-Instruct (123B) (dense instruction-tuned model).
Throughout we report exactly one checkpoint per family (never both base \emph{and} instruct/chat), and we interpret results in light of the dense-versus-MoE distinction (all-parameter vs sparsely activated compute) and the growing class of \emph{reasoning} models that trade extra train-time or test-time compute for stronger multi-step problem solving.

Working with large language models (LLMs) always involves prompts.
Two strategies are natural: (i) optimise a separate prompt per model, or (ii) fix a single prompt that remains unchanged across models and runs.
We adopt the latter for three reasons: \emph{comparability}, \emph{reproducibility}, and \emph{engineering realism}.
Per-model prompt tuning risks cherry-picking idiosyncratic templates that inflate best-case performance but obscure genuine capability differences; it also introduces substantial researcher degrees of freedom.
Nevertheless using a single template has an obvious downside: It likely underestimates each model's best-case performance under bespoke prompt engineering.
We accept this, because our goal is \emph{relative} capability under a standardised interface. 

Our prompt design is illustrated in Figure \ref{fig:Prompt:Navigation:Zero:Shot}. 
This is used as a \emph{base template} with optional few-shot extension.
The difference between \textit{Navigation} and \textit{Exploration} is kept as small as possible and is illustrated in Figure \ref{fig:Prompt:Exploration:Zero:Shot}. 
The template has four blocks:

\begin{enumerate}
    \item Role \& task: one paragraph that frames the agent as a navigation module with the optimisation objective (short path to \texttt{g}) and a hard safety constraint (never step onto \texttt{W}).
    \item Gridworld reference: a compact glossary (\texttt{W}, \texttt{.}, \texttt{g}, \texttt{?}, \texttt{R}) and our coordinate convention (row--major, origin at top left). Re-stating these per turn reduces drift across long sequences.
    \item Action schema \& output contract: a small table mapping directions to integers, plus a \emph{strict} JSON payload wrapped in sentinel tags. Redundant fields (\texttt{direction\_str} \emph{and} \texttt{direction\_int}) allow a simple consistency check and catch off-by-one errors.
    \item Current status: a concise header with (prev~$\rightarrow$~curr) move, last action, reward note, and the current ASCII grid. We intentionally provide only the immediately preceding grid to cap context length and avoid ``lost in the middle'' effects when appending many snapshots.
\end{enumerate}

\begin{figure}[htb]
    \centering
\begin{Verbatim}
# Exploration – Zero Shot

## Task

You are the robot's navigation module. At each iteration, decide a single cardinal 
move that will safely uncover new territory, avoid collisions with walls (W), and 
maximise the number of discovered grid tiles. Continue until every "?" tile has been 
observed.

## Gridworld Reference

[...]
\end{Verbatim}
    \caption{Task description in a prompt for Exploration Zero-Shot (Excerpt)}
    \label{fig:Prompt:Exploration:Zero:Shot}
\end{figure}

\begin{figure}[hptb]
    \centering
\begin{Verbatim}
# Navigation – Zero-Shot

## Task

You are the robot's navigation module.
At each iteration, decide a single cardinal move that will safely move 
the robot on the shortest path towards the goal position.
Do not choose an action that will move the robot on a wall tile; 
under no circumstances.

## Gridworld Reference

'W': Wall (impassable)
'.' or ' ': Free cell
'g': Goal for navigation tasks
'?': Unexplored tile for exploration tasks
'R': Robot’s current position

*Coordinates* '(row, col)' with origin **(0, 0)** at the **top-left** corner.

Each move unveils a 5x5 grid around the robot if it is not already discovered. As an
observation you get the grid world with all unexplored tiles and all the revealed
tiles with the current position of the robot marked with "R".

## Actions

| Action | Move  | row | col |
| ------ | ----- | --- | --  |
| '0'    | Down  | +1  | 0   |
| '1'    | Right | 0   | +1  |
| '2'    | Up    | –1  | 0   |
| '3'    | Left  | 0   | –1  |

**Never** choose an action that would put the robot on a space with 'W'.

## Output

Output the result in the following format:
<output>{"direction_str": "<direction>", "direction_int": <int>}</output>.
The valid mappings are: down -> 0, right -> 1, up -> 2, left -> 3.
No other output is allowed.
Example: <output>{"direction_str": "down", "direction_int": 0}</output>

## Current Status

### Iteration 9

- prev → curr  : [[14  5]] -> [[13  5]]  
- action       : (2, 'up')  
- reward       : -0.01: No note.

<observation>
['????????WWWWW????????'
 '????????..g..????????'
 '????????.....????????'
 '????????.....????????'
 '?????????????????????'
 '?????????????????????'
 '?????????????????????'
 '?????????????????????'
 '?????????????????????'
 '?????????????????????'
 '?????????????????????'
 '???...WW?????????????'
 '???...WWWWWWW????????'
 '???..R.......????????'
 '???..........????????'
 '???..........????????'
 '???..........????????'
 '????????.....????????'
 '????????WWWWW????????']
</observation>
\end{Verbatim}
    \caption{Example prompt for Navigation Zero-Shot in \emph{Medium} environment}
    \label{fig:Prompt:Navigation:Zero:Shot}
\end{figure}

This structure emphasises \emph{minimalism for reliability}. A short role paragraph plus a tabular action schema reduces paraphrase opportunities and anchors the output space to four symbols.
The \emph{closed-form output} 
allows trivial parsing with a single regular expression, tolerates minor surrounding chatter and enables automated rejection/retry without side effects.
By providing \emph{local observability with a global intent}, i.e.\ we state the shortest-path objective yet expose only local information (a $5\times5$ reveal and the current global grid), we discourage speculation about unseen structure while still biasing the model towards corridor-following.
This compact-history design is an explicit \emph{mitigation against long-context brittleness} ("lost in the middle"/ context rot), preserving the decision-relevant state while avoiding distractors from redundant map snapshots. 
Longer histories empirically caused looping and distraction.
Thus trading trajectory completeness for robustness.

Unless restricted by the vendor API, we fix decoding hyperparameters across all models to a common default (temperature=0.8, top\_k=40, top\_p=0.9).
Access is exclusively programmatic, either via a local Ollama runtime or via vendor HTTP APIs and no browser UI is used.
For the proprietary GPT-5 and GPT-5-mini these controls are unavailable or are ignored on an API level. Instead, the API exposes a coarse \texttt{reasoning.effort} setting {\textit{low}, \textit{medium}, \textit{high}}.
We keep \texttt{reasoning.effort} at its default (\textit{medium}) and do not vary it. 
This breaks strict parameter parity and may understate those models' best-case performance under bespoke decoding.
We accept this limitation on the individual capability ceiling for the same reasons as above: \emph{comparability}, \emph{reproducibility}, and \emph{engineering realism at the API level}. Our aim is \emph{relative} capability under a standardised programmatic interface, not per-model best-case performance via private or unavailable hyperparameters.

We compare two prompting regimes using the same base template. In zero-shot runs we present only the four instruction blocks above.
In the five-shot run we append the base template with five in-context examples that demonstrate valid and invalid moves: each of the four robot movements (down/right/up/left) and a collision with a wall.
In five-shot runs these five examples are prepended as full blocks (observation before action, chosen action, observation after action), so the model sees concrete demonstrations of revealing new grid cells, the effect of partial observability on directional outcomes, and collisions that leave the observation unchanged.

\section{Results and Discussion}
\label{Results:and:Discussion}

We deliberately do not include classical online re-planners as baselines; instead, we bracket performance between a \emph{lower bound} (uniform random action policy) and an \emph{upper bound} (oracle metrics assuming full map knowledge: maximum coverable cells for coverage and the shortest path length for navigation).
This keeps our evaluation aligned with the strict no-tool constraint and isolates text-only action selection under partial maps. 

\begingroup
\renewcommand{\arraystretch}{1.1}
\newcolumntype{C}{>{\hspace{4pt}}c<{\hspace{4pt}}}

\begin{table}[t]
\centering
\begin{tabular}{|l|C C|C C|}
\hline
 & \multicolumn{2}{C|}{Exploration}  & \multicolumn{2}{C|}{Navigation}\\
 Model & 0 - Shot & 5 - Shot & 0 - Shot & 5 - Shot \\ \hline
Random Actions & \multicolumn{2}{C|}{53.71\,\%} & \multicolumn{2}{C|}{8.00\,\% (282)} \\ \hline
GPT-5 (Reasoning) & 100.00\,\% & 100.00\,\% & 100.00\,\% (26) & 100.00\,\% (26) \\
GPT-OSS-120B (Reasoning) & {100.00\,\% }& {100.00\,\% } & 100.00\,\% (29) & 100.00\,\% (28)\\
GPT-5-mini (Reasoning) & 99.47\,\% & 100.00\,\% & 100.00\,\% (60) & 100.00\,\% (28)\\
Gemma-3-27B-Instruct & \textcolor{red}{39.97\,\%} & 61.23\,\% & 100.00\,\% (106) & 100.00\,\% (37)\\
Llama-3.3-70B-Inst. & \textcolor{red}{34.85\,\%} & \textcolor{red}{40.96\,\%} & 100.00\,\% (60) & 38.46\,\% (91)\\
DeepSeek-R1 (685B / 37B) & \textcolor{red}{43.98\,\%} & \textcolor{red}{37.65\,\%} & 30.00\,\% (189) & 30.77\,\% (160)\\
Mistral-Large-Inst. (123B) & \textcolor{red}{30.34\,\%} & \textcolor{red}{37.09\,\%} & \textcolor{red}{0.00\,\% ($\infty$)} & 23.08\,\% (149)\\
Qwen3-235B (Reasoning) & \textcolor{red}{52.73\,\%} & 57.62\,\% & \textcolor{red}{0.00\,\% ($\infty$)} & 7.69\,\% (93)\\
Qwen3-QwQ-32B & \textcolor{red}{43.16\,\%} & 66.37\,\% & \textcolor{red}{0.00\,\% ($\infty$)} & 7.69\,\% (195)\\
\hline
\end{tabular}
\caption{Easy: Per model exploration coverage for exploration and success rate (average steps) for navigation. The oracle path is 26 cells. $n=13$}
\label{tab:easy}
\end{table}
\endgroup
\begingroup
\renewcommand{\arraystretch}{1.1}
\newcolumntype{C}{>{\hspace{4pt}}c<{\hspace{4pt}}}
\begin{table}[t]
\centering
\begin{tabular}{|l|C C|C C|}
\hline
 & \multicolumn{2}{C|}{Exploration}  & \multicolumn{2}{C|}{Navigation}\\
 Model & 0 - Shot & 5 - Shot & 0 - Shot & 5 - Shot \\ \hline
Random Actions & \multicolumn{2}{C|}{45.76\,\%} & \multicolumn{2}{C|}{3.00\,\% (316)} \\ \hline
GPT-5 (Reasoning)& 100.00\,\% & 100.00\,\% & 100.00\,\% (48) & 100.00\,\% (44)\\
GPT-OSS-120B (Reasoning)& 99.50\,\% & 99.94\,\% & 100.00\,\% (52) & 100.00\,\% (46)\\
GPT-5-mini (Reasoning)& 99.00\,\% & 100.00\,\%& 0.00\,\% ($\infty$) & 0.00\,\% ($\infty$)\\
Qwen3-QwQ-32B & \textcolor{red}{22.70\,\%} & 68.22\,\% & \textcolor{red}{0.00\,\% ($\infty$)} & 30.77\,\% (122)\\
DeepSeek-R1 (685B / 37B) & \textcolor{red}{42.17\,\%} & \textcolor{red}{45.59\,\%} & \textcolor{red}{0.00\,\% ($\infty$)} & 7.69\,\% (183)\\
\hline
Mistral-Large-Inst. (123B) & \textcolor{red}{29.16\,\%} & \textcolor{red}{39.98\,\%} & \textcolor{red}{0.00\,\% ($\infty$)} & \textcolor{red}{0.00\,\% ($\infty$)}\\
Gemma-3-27B-Instruct & \textcolor{red}{17.26\,\%} & \textcolor{red}{43.63\,\%} & \textcolor{red}{0.00\,\% ($\infty$)} & \textcolor{red}{0.00\,\% ($\infty$)}\\
Qwen3-235B (Reasoning) & \textcolor{red}{23.04\,\%} & \textcolor{red}{24.40\,\%} & \textcolor{red}{0.00\,\% ($\infty$)} & \textcolor{red}{0.00\,\% ($\infty$)}\\
Llama-3.3-70B-Inst. & \textcolor{red}{16.09\,\%} & \textcolor{red}{18.92\,\%} & \textcolor{red}{0.00\,\% ($\infty$)} & \textcolor{red}{0.00\,\% ($\infty$)} \\
\hline
\end{tabular}
\caption{Medium: Per model exploration coverage for exploration and success rate (average steps) for navigation. The oracle path is 25 cells. $n=13$}
\label{tab:medium}
\end{table}
\endgroup
\begingroup
\renewcommand{\arraystretch}{1.1}
\newcolumntype{C}{>{\hspace{4pt}}c<{\hspace{4pt}}}
\begin{table}[t]
\centering
\begin{tabular}{|l|C C|C C|}
\hline
 & \multicolumn{2}{C|}{Exploration}  & \multicolumn{2}{C|}{Navigation}\\
 Model & 0 - Shot & 5 - Shot & 0 - Shot & 5 - Shot \\ \hline
Random Actions & \multicolumn{2}{C|}{35.64\,\%} & \multicolumn{2}{C|}{7.80\,\% (283)} \\ \hline
GPT-5 (Reasoning) & 89.22\,\% & 100.00\,\% & 100.00\,\% (32) & 100.00\,\% (32)\\
GPT-5-mini (Reasoning) & 62.91\,\% & 83.21\,\% & 100.00\,\% (56) & 100.00\,\% (50)\\
GPT-OSS-120B (Reasoning) & 58.56\,\% & 45.21\,\% & 100.00\,\% (43) & 100.00\,\% (40)\\
DeepSeek-R1 (685B / 37B) & 42.97\,\% & 51.55\,\% & 30.00\,\% (207) & 53.85\,\% (123)\\
Qwen3-QwQ-32B & \textcolor{red}{25.97\,\%} & 41.75\,\% & \textcolor{red}{0.00\,\% ($\infty$)} & 15.38\,\% (114)\\
Mistral-Large-Inst. (123B) & \textcolor{red}{28.89\,\%} & 40.55\,\% & \textcolor{red}{0.00\,\% ($\infty$)} & 15.38\,\% (168)\\
Gemma-3-27B-Instruct& \textcolor{red}{21.25\,\%} & 38.44\,\% & \textcolor{red}{0.00\,\% ($\infty$)} & \textcolor{red}{0.00\,\% ($\infty$)}\\
\hline
Llama-3.3-70B-Inst. & \textcolor{red}{16.04\,\%} & \textcolor{red}{21.08\,\%} & \textcolor{red}{0.00\,\% ($\infty$)} & \textcolor{red}{0.00\,\% ($\infty$)}\\
Qwen3-235B (Reasoning) & \textcolor{red}{29.11\,\%} & \textcolor{red}{29.62\,\%} & \textcolor{red}{0.00\,\%  ($\infty$)} & \textcolor{red}{0.00\,\% ($\infty$)}\\
\hline
\end{tabular}
\caption{Hard: Per model exploration coverage for exploration and success rate (average steps) for navigation. The oracle path is 24 cells. $n=13$}
\label{tab:hard}
\end{table}
\endgroup

Across models and environments we observe a consistent UP/RIGHT action bias, either as a direct sequence of actions or a more general indication of direction over longer sequences. 
It is most pronounced in models that underperform the random baseline and the \emph{Instruct} models, especially on the \emph{exploration} task where no explicit target position is given.
Reasoning-tuned models are less affected with the only exception being the Qwen3-235B model (s. Table~\ref{tab:easy} and Table~\ref{tab:medium}). 
With this action bias the models often fall into ``death loops'' that repeat for tens to $>100$ steps. 
The UP/RIGHT prior makes corners with walls above and to the right especially problematic.
In the \emph{Easy} environment this frequently traps agents in the upper-right corner, yielding $\sim30\,\%$\,--\,$45\,\%$ exploration coverage.
In the \emph{Medium} environment the first such a trap (where the right obstacle meets the outer wall) yields $\sim16\,\%$\,--\,$25\,\%$ exploration coverage, and a second \emph{honeypot} near the upper-left obstacle explains runs stalling at $\sim40\,\%$\,--\,$50\,\%$ exploration coverage before later getting stuck again in the upper-right corner. 
The \emph{Hard} environment contains many such right-angled stalling points where several models plateau around $\sim16$\,--\,$29\%$ exploration coverage.
Additionally some models show central-column looping, where they cannot leave the central area of row three to five. These models stall at $\sim40\,\%$\,--\,$50\,\%$ exploration coverage.

Five-shot tasks generally mitigate the RIGHT bias and reduce invalid moves, but the effect is model-dependent: for Llama-3.3-70B-Instruct, adding examples markedly degrades Navigation versus zero-shot, suggesting that a minimally constrained prompt can be preferable. 
Finally, the ordering in \emph{Medium} (Table~\ref{tab:medium}) should not be read as a strict ranking: GPT-5-mini fails on Navigation while Qwen3-QwQ-32B is notably weaker on Exploration, placing them on opposite sides of the trade-off. 
In the \emph{Hard} environment, many models prematurely commit to moving straight up through the central columns, because under the $5\times5$ view it is plausibly the shortest path to the goal but it divergent from the oracle path.
Thus they can never reach the truly shortest path with exploring the rest of the environment.
It seems our few-shot runs are consistent with \cite{wei22cot}: for reasoning-tuned models, $n$-shot behaves like a compute switch, markedly reducing invalid moves and shortening paths (e.g., GPT-5-mini drops from 60 to 28 steps on \emph{Easy}; Table~\ref{tab:easy}), whereas dense Instruct baselines exhibit little systematic gain.
This supports the view that CoT/few-shot prompting can be helpful yet remains format-sensitive and does not substitute for robust long-horizon planning under partial maps.

In line with real-life robotics applications as a secondary objective the number of Wall collisions should be kept to ideally zero.
However, for most open-weight models the number of collisions could not be eliminated during exploration and navigation tasks only decrease with five-shot examples. 
The only zero and near-zero collision rates appear during the navigation task for the open-weight model GPT-OSS-120B and the closed-weight models GPT-5, GPT-5-mini. 

Across the three environments, the clearest regularity is that \emph{reasoning-tuned} models are the only ones that work well on \emph{Navigation}, especially under five-shot prompting.
Systems explicitly trained to ``think before answering'' -- e.g., RL- or search-augmented reasoners -- benefit disproportionately from examples, while classic dense \emph{Instruct} chat models (even at 70--123B parameters) remain near-zero on Navigation despite competent Exploration.
A second, practically salient pattern is that \emph{active size per token} matters more than total parameter count: modern Mix of Experts (MoE) models expose huge totals but activate a small subset of experts per token; in the results, performance tracks the training pipeline and test-time reasoning budget far more than raw scale.
Third, few-shot prompting behaves like a compute ``switch'': it activates planning/search in reasoners but yields little lift for dense Instruct models.
Fourth, intra-family contrasts are visible: the dense, dedicated reasoner Qwen3-QwQ-32B often matches or exceeds the larger MoE Qwen3-235B on Navigation, consistent with either (i) better specialisation of the SFT/RL recipe for algorithmic search or (ii) incomplete triggering of a model's internal ``thinking mode'' by the prompt template.
Taken together, the models cluster into three groups: (A) dedicated reasoners (convert examples into long-horizon control, robust to environment difficulty), (B) classic dense Instruct chat models (strong on dialogue/knowledge but weak on planning without extra search), and (C) hybrids with explicit mode control whose outcomes hinge on reliably engaging the thinking pathway.
Overall, \emph{training regimen + test-time compute} are better predictors of Navigation than either total or active parameter counts.
Beyond this it is plausible that the OpenAI models GPT-OSS-120B, GPT-5 and GPT-5-mini in this set have seen related tasks during training.

Viewed against prior work from section \ref{sec:sota} on text-based planning, our results call for a qualified reading.
For classical instruct models (typical up to 2024), navigation degrades with longer horizons and partial observability, often failing outright, in line with \cite{huang22,kim24textgrid}. 
By contrast, recent reasoning-tuned systems (notably GPT-5 and GPT-OSS-120B) shift the failure mode from \emph{success} to \emph{efficiency}: they reach the goal reliably across all three layouts (100\,\% success; Tables~\ref{tab:easy}--\ref{tab:hard}) yet incur substantial path overhead relative to the oracle, especially in \emph{Medium} (e.g., GPT-5: 48/44 steps vs.\ $24$; GPT-OSS-120B: 52/46 vs.\ $24$; Table~\ref{tab:medium}) and still noticeable in \emph{Hard} (GPT-5: 32 vs.\ $26$; GPT-OSS-120B: 43/40 vs.\ $26$; Table~\ref{tab:hard}), while \emph{Easy} includes an optimal run (GPT-5: 26; Table~\ref{tab:easy}). 
Trajectories exhibit coarse, model-agnostic heuristics -- early straight-line commitment and a mild \texttt{UP}/\texttt{RIGHT} preference -- that are locally rational under a $5\times5$ view but globally suboptimal once branches and cul-de-sacs unfold; this representation-induced behaviour matches the sensitivity reported for textual encodings \cite{martorell25text2space}.
Few-shot demonstrations help primarily for reasoners, like reducing invalid moves and shortening paths, yet provide limited and format-sensitive gains for dense instruct models, consistent with the mixed efficacy of CoT/few-shot prompting \cite{wei22cot}. 
Overall, our findings corroborate that prompt-only next-action control remains brittle as global structure increases \cite{huang22,kim24textgrid}, but they also indicate that current reasoning-tuned models have internalised workable \emph{heuristic} navigation strategies that deliver robustness on success at the expense of efficiency.
The persistent gap to oracle paths points toward explicit planning or hybridisation -- e.g., lightweight frontier selection or incremental shortest-path search -- which prior LLM+tool frameworks already suggest can reduce long-horizon detours \cite{yao22react,li2025gridroute}, in a division of labour akin to language for high-level intent with classical control for low-level execution \cite{ahn22saycan}.
Finally, the uneven difficulty profile across layouts (e.g., GPT-5-mini failing on \emph{Medium} but not on \emph{Hard}) hints at limited systematic generalisation over structural combinations rather than fully compositional planning rules, partially echoing observations from benchmarks \cite{kim2020cogs}.

\section{Conclusion and Future Prospects}
\label{sec:Conclusion:and:Future:Prospects}
One of our initial questions was: to what extent can LLMs act as controllers when the environment is only partially observed? Our results show that reasoning-tuned LLMs can reliably provide task-level control policies via a text-only interface, achieving consistent goal completion across layouts even when tool use and program synthesis are prohibited.
However, they act through heuristics rather than algorithmic search, which leaves a persistent efficiency gap to oracle paths and exposes representation-sensitive behaviours.
Crucially, training regimen and test-time deliberation predict navigation ability better than raw parameter count.
There is, nonetheless, a clear positive trajectory: when models are fine-tuned for control, low-level behaviour improves (fewer invalid moves, more stable rollouts), and -- despite the strict no-code/no-tool constraint -- several systems performed surprisingly well, notably the open-weight GPT-OSS-120B, indicating that progress is not confined to proprietary models.
From a Green AI perspective, LLM inference is more energy-intensive than classical online planning; hence LLM calls are best gated to infrequent, abstract decision points, with high-rate control and path optimisation left to classical algorithms.
To justify their energy and carbon cost at scale, LLM controllers must ultimately outperform classical baselines -- or unlock capabilities that classical stacks cannot provide.
In a robotics stack, LLMs are most effective at the task/behaviour layer -- interpreting goals, prioritising frontiers, and revising local intent -- while online re-planners handle path optimisation and certified controllers govern low-level actuation and safety.
Near-term gains will come from lightweight hybridisation and simple hierarchical designs that retain the strict text interface but pair the LLM policy with classical planning.
In short: LLMs can act as controllers to the extent that they deliver robust high-level policies; efficiency and guarantees remain with classical planning and verified control -- yet the current trend suggests this boundary is steadily shifting.

\section*{Acknowledgements}

Part of this work was funded by the Federal Ministry for Economic Affairs and Energy. FKZ: 16TNWB0024C - TrAIBeR.NRW

\end{document}